# An MLI-Guided Framework for Subgroup-Aware Modeling in Electronic Health Records (AdaptHetero)

Ling Liao[1, 2], Eva Aagaard[3]

1. Computational and Systems Biology, Washington University, MO, U.S.A, 63130
2. Biomedical Deep Learning LLC, MO, U.S.A, 63130
3. Department of Medicine, School of Medicine, Washington University, MO, U.S.A, 63130

Correspondence: Ling Liao (lingliao@wustl.edu), Eva Aagaard (aagaarde@wustl.edu)

**Abstract**
Machine learning interpretation (MLI) has primarily been leveraged to foster clinician trust and extract insights from electronic health records (EHRs), rather than to guide subgroup-specific, operationalizable modeling strategies. To bridge this gap, we propose AdaptHetero, a novel MLI-driven framework that transforms interpretability insights into actionable guidance for tailoring model training and evaluation across subpopulations. Evaluated on three large-scale EHR datasets—GOSSIS-1-eICU, WiDS, and MIMIC-IV—AdaptHetero consistently uncovers heterogeneous model behaviors in predicting ICU mortality, in-hospital death, and hidden hypoxemia. Integrating SHAP-based interpretation with unsupervised clustering, AdaptHetero identifies clinically meaningful, subgroup-specific characteristics, improving predictive performance across many subpopulations (with gains up to 174.39%) while proactively flagging potential risks in others. These results highlight the framework's promise for more robust, equitable, and context-aware clinical deployment.

**Introduction**
Machine learning interpretation (MLI) techniques are increasingly leveraged in the analysis of electronic health records (EHRs) to reveal latent clinical patterns and support trustworthy decision-making in high-stakes healthcare settings.[1,2,3,4] However, even when interpretation methods consistently identify similar top predictors (e.g., lactate level, ventilated status for intensive care unit (ICU) mortality), predictive performance can vary substantially across hospitals—or even within the same hospital system—due to the inherent heterogeneity of EHR data.[1,5,6,7] This variability poses a fundamental challenge to the deployment and generalizability of ML models in clinical practice, perpetuating disparities by delivering suboptimal predictions for some patient groups and potentially harmful, misleading outputs for others.

To address this issue, current strategies primarily rely on multi-site training paradigms, including federated learning, rigorous external validation, and deep learning-based representation learning.[8,9,10,11,12,13,14] While these methods have yielded promising preliminary results, they fall short of fully overcoming critical issues like data and semantic heterogeneity, restricted access to external validation datasets, high computational overhead.[15,16,17] Consequently, current approaches provide only partial solutions, with model robustness and generalizability still far from clinical readiness. As a result, only a handful of ML models have been deployed in clinical practice, whereas many more are available online but not yet clinically utilized.

We propose AdaptHetero, a novel MLI-guided framework that leverages lightweight learning to transform model interpretation outputs into actionable guidance for subgroup-aware model training and evaluation (see Fig. 1). AdaptHetero establishes a closed-loop paradigm: initial subgrouping is informed by SHAP(SHapley Additive exPlanations)[18]-based interpretation, and the interpretability results of trained subgroup-specific models are then used to assess the consistency and distinctiveness of feature attributions across subgroups. We validate AdaptHetero on three large-scale datasets— GOSSIS-1-eICU[13], WiDS[19], and MIMIC-IV[20]—and demonstrate that it consistently identifies subgroup-specific performance disparities and, in many cases, outperforms baseline models trained on pooled data, while also flagging potential risks in others.

**Fig. 1** Outlined AdaptHetero framework architecture. Input data are split into training and testing sets. After initial model training, patient subgroups are derived from the training data by integrating SHAP-based interpretation with unsupervised clustering. These cluster assignments are then propagated to the testing data. Subsequently, models are retrained within each subgroup using the corresponding training data and evaluated on the matching test data. Following retraining, subgroup-specific SHAP values are recomputed on the training sets and compared with initial interpretations to iteratively refine clustering quality and enhance predictive performance, thereby forming a closed-loop feedback system. Finally, predictive performance is assessed on the testing data using AUPRC and log loss metrics, while SHAP representations are analyzed to validate model behavior and heterogeneity across subpopulations.

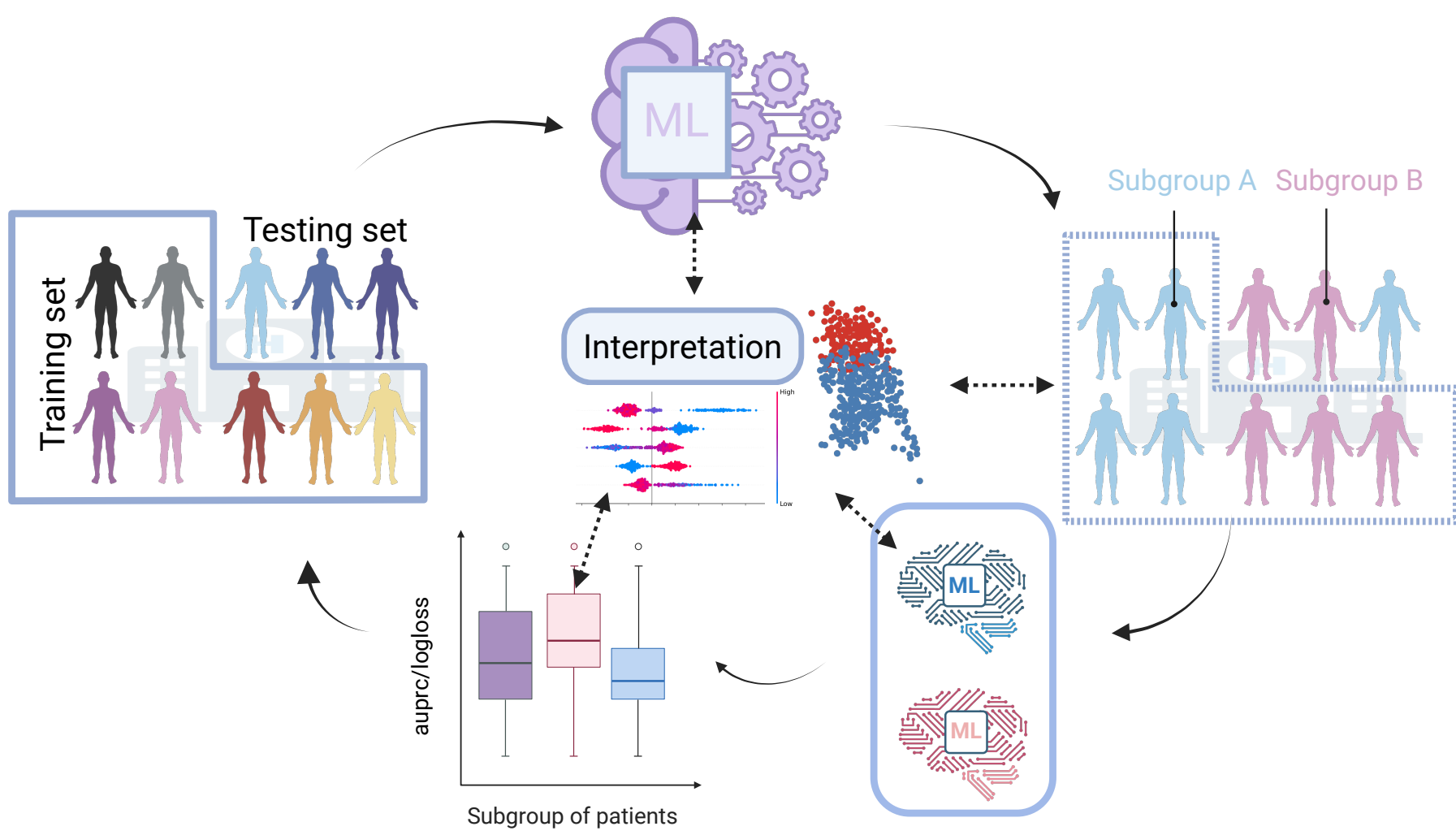

## Results

### Identifying heterogeneity and subgroups in hospital systems

We first evaluated AdaptHetero's ability to uncover previously unidentified patient subgroups based on model interpretation patterns derived from EHR data. To this end, we trained an XGBoost (Extreme Gradient Boosting) model on 70% of each cohort and tested it on the remaining 30%. SHAP values were computed for both training and test sets. The SHAP values from the training set were then used to perform unsupervised clustering via HDBSCAN (Hierarchical Density-Based Spatial Clustering of Applications with Noise)[21,22]. The resulting clusters, visualized in Figs. 2a–2e, represent distinct patient subpopulations characterized by heterogeneous model behavior in the training data (corresponding to hospitals 73, 458, and 420 in the eICU dataset for mortality prediction; hospital 19 in the WiDS dataset for in-hospital death prediction; and the MIMIC-IV dataset for hidden hypoxemia prediction).

We propagated the discovered clusters to the held-out test set's SHAP-derived feature representations using an unsupervised algorithm K-nearest neighbors (KNN)[23,24]. Finalized subgroup inclusion, based on clustering, are detailed in Supplementary Table 1 (This step was conducted iteratively and confirmed by examining heterogeneity in SHAP-based feature representations within the training sets.). We then evaluated the performance of the trained XGBoost model on the test set within each subgroup, comparing the overall test cohort ("All") with Subgroup A and Subgroup B, respectively (Fig. 2f, 2g; legend in Fig. 2h).The performance was quantified using the AUPRC ( Area Under the Precision–Recall Curve ), chosen for its robustness to class imbalance and its emphasis on accurately identifying positive cases— often clinically critical in EHR data[1]. Statistical significance was assessed via the Mann–Whitney U test, a non-parametric method well-suited for the limited number of AUPRC observations.[1,25] In Figs. 2f–2g, we observed statistically significant differences in predictive performance (AUPRC) and probability calibration (Logloss) between all the identified subgroups A and B.

Collectively, these results highlight AdaptHetero's capability to reveal meaningful heterogeneity within EHR data, evidenced by significant variations in model performance across patient subpopulations, representing an important step toward advancing personalized, equitable medicine in high-stakes clinical settings.

**Fig. 2** AdaptHetero-derived patient subgroups based on SHAP clustering and their impact on predictive modeling performance. **a–e** Scatter plots of training samples from hospital 73, 458, and 420 (eICU data), hospital 19 (WiDS data), and MIMIC IV data, projected into two dimensions using UMAP, with points colored according to HDBSCAN cluster assignments. **f-g** Performance of clinical outcome predictions on the eICU, WiDS, and MIMIC-IV data, which was evaluated on a bootstrap-sampled dataset with median values representing the central tendency of model metrics across resamples. **h** detailed hospital information for **f** and **g**. Statistical significance of differences in test set between Subgroups and correspondingly retrained Subgroups were assessed using the non-parametric Mann–Whitney U test. * , **, and *** denote p values <0.05, <0.01, and <0.001, respectively. **h**. Legends of the datasets and prediction goals for 2f and 2g.

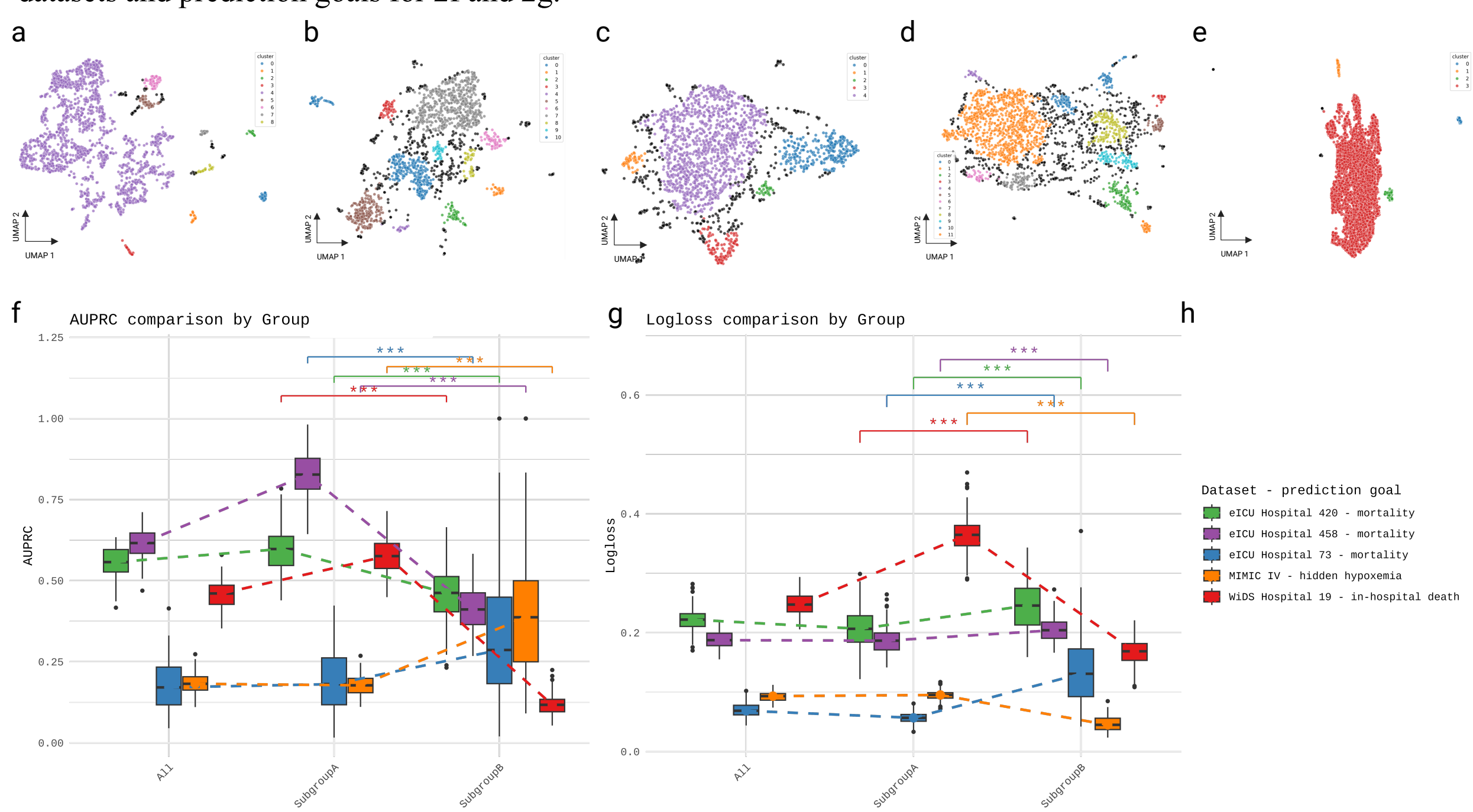

**Retraining model on subgroups**

To assess whether subgroup-specific modeling improves predictive performance, we retrained XGBoost models separately for Subgroup A and Subgroup B using only their respective training samples. We then evaluated these models on their corresponding test subsets and compared the results against baseline: (i) the global model's performance on the same subgroup test data. We included a control comparison as well (ii): the global model's performance on the subgroup test data versus the entire test set. Results are summarized in Table 1.

In table 1, median AUPRC values and percentage changes are reported for each comparison. Statistically significant differences (Mann–Whitney U test, $p < 0.05$) are noted. By comparing with baseline i, for example, retraining the Subgroup A model notably improved mortality prediction on eICU Hospital 73 (from 0.18 to 0.26; +42.39%), and retraining the Subgroup B model enhanced hidden hypoxemia prediction on MIMIC IV (from 0.39 to 0.5; +29.09%), both statistically significant.

Examining the control comparison (ii), the AUPRC increased modestly from 0.17 to 0.18 (+5.99%) for mortality prediction on eICU Hospital 73 between the full test set and Subgroup A, and substantially from 0.18 to 0.39 (+112.57%) for hidden hypoxemia prediction on MIMIC IV between the full test set and Subgroup B.

Collectively, these examples suggest that evaluating models on subgroup-specific test sets can yield notable performance benefits, which can be further enhanced by retraining models exclusively on those subpopulations. Specifically, for hidden hypoxemia prediction on MIMIC IV data, the AUPRC improved from 0.18 (full test set) to 0.39 (Subgroup B), and further increased to 0.50 after retraining on Subgroup B, representing a cumulative gain of +174.39%.

Conversely, retraining the Subgroup A model for mortality prediction on eICU Hospital 420 resulted in a minor, non-significant decrease (from 0.60 to 0.59; -1.31%), while retraining the Subgroup B model on eICU Hospital 73 produced a significant decline (from 0.29 to 0.09; -67.92%).

Nonetheless, when considering control comparison (ii), mortality prediction AUPRC was higher in Subgroup A than the full test set on eICU Hospital 420 (0.60 vs. 0.56; +7.23%), and similarly elevated in Subgroup B compared to the full set on eICU Hospital 73 (0.29 vs. 0.17; +67.44%). These results collectively indicate that it is advisable to continue using the globally trained model, while evaluating performance on relevant subgroups rather than the entire test set —so as not to obscure potential subgroup-specific risks.

Taken all together, findings in this section support that subgroup-specific modeling can not only boost predictive performance but also pave the way toward precision medicine by supporting tailored model development and selection.

**Table 1.** Comparison of Median AUPRC and Statistical Significance Across Subgroups, Datasets, and Prediction Goals. **Dataset - prediction** Dataset and associated prediction task used for the comparison. **comparison** Groups being compared; statistical significance is computed between them. **median_group1, median_group2** Median AUPRC for each group listed in the comparison column. **Pct_change** Percentage change relative to median_group1,

calculated as (median_group2 - median_group1) / median_group1 × 100%. **significance** Indicates whether the difference is statistically significant (Mann–Whitney U test, $p < 0.05$) [1] Rounded to two decimal places for display. [2] Calculated using unrounded median values, then rounded to two decimal places for display.

| Dataset - prediction | comparison | median_group1[1] | median_group2[1] | pct_change[2] | significance |
|---|---|---|---|---|---|
| MIMIC IV - hidden hypoxemia | All vs SubgroupB | 0.18 | 0.39 | +112.57% | TRUE |
| eICU Hospital 73 - mortality | All vs SubgroupB | 0.17 | 0.29 | +67.44% | TRUE |
| eICU Hospital 73 - mortality | SubgroupA vs SubgroupA_retrained | 0.18 | 0.26 | +42.39% | TRUE |
| eICU Hospital 458 - mortality | All vs SubgroupA | 0.62 | 0.83 | +34.27% | TRUE |
| MIMIC IV - hidden hypoxemia | SubgroupB vs SubgroupB_retrained | 0.39 | 0.5 | +29.09% | TRUE |
| WiDS Hospital 19 - in-hospital death | All vs SubgroupA | 0.46 | 0.58 | +24.99% | TRUE |
| eICU Hospital 420 - mortality | All vs SubgroupA | 0.56 | 0.6 | +7.23% | TRUE |
| eICU Hospital 73 - mortality | All vs SubgroupA | 0.17 | 0.18 | +5.99% | FALSE |
| MIMIC IV - hidden hypoxemia | SubgroupA vs SubgroupA_retrained | 0.18 | 0.19 | +5.42% | TRUE |
| WiDS Hospital 19 - in-hospital death | SubgroupA vs SubgroupA_retrained | 0.58 | 0.6 | +4.96% | TRUE |
| eICU Hospital 458 - mortality | SubgroupA vs SubgroupA_retrained | 0.83 | 0.85 | +2.71% | TRUE |
| eICU Hospital 420 - mortality | SubgroupA vs SubgroupA_retrained | 0.6 | 0.59 | -1.31% | FALSE |
| MIMIC IV - hidden hypoxemia | All vs SubgroupA | 0.18 | 0.18 | -2.57% | FALSE |
| eICU Hospital 458 - mortality | SubgroupB vs SubgroupB_retrained | 0.41 | 0.35 | -14.37% | TRUE |
| eICU Hospital 420 - mortality | All vs SubgroupB | 0.56 | 0.46 | -17.05% | TRUE |
| WiDS Hospital 19 - in-hospital death | SubgroupB vs SubgroupB_retrained | 0.12 | 0.1 | -17.89% | TRUE |
| eICU Hospital 420 - mortality | SubgroupB vs SubgroupB_retrained | 0.46 | 0.38 | -18.42% | TRUE |
| eICU Hospital 458 - mortality | All vs SubgroupB | 0.62 | 0.41 | -33.22% | TRUE |
| eICU Hospital 73 - mortality | SubgroupB vs SubgroupB_retrained | 0.29 | 0.09 | -67.92% | TRUE |
| WiDS Hospital 19 - in-hospital death | All vs SubgroupB | 0.46 | 0.12 | -74.50% | TRUE |

### Validating clinical heterogeneity on test data

Upon completing the closed-loop, we analyzed feature attributions derived from SHAP values within the test data for each identified subgroup. Specifically, SHAP values were calculated for all test instances, as well as separately for Subgroup A, Subgroup B, and their corresponding retrained models.

As illustrated in Fig. 3, while most features were frequently reported as important predictors within their respective clinical contexts (e.g., the motor component of the Glasgow Coma Scale as a key neurological indicator of ICU mortality risk, and peripheral oxygen saturation as a proxy for respiratory compromise), notable divergences were observed between Subgroup A and Subgroup B in both the types of top-ranked features and their relative importance (see Figs. 3f–3j and 3k–3o). An exception was the contribution of missing values, which appeared among the top features in some groups (see Figs. 3a, 3f, 3k, 3p, 3v), despite being typically removed or imputed prior to model training. As expected, retraining models introduced further variations in feature attribution (see Figs. 3p–3t and 3u–3y).

Intriguingly, when we jointly considered the SHAP-based feature representations across the test groups (All, Subgroup A, Subgroup B, and their respective retrained models) along with the corresponding median AUPRC values (Table 1), we observed a clear correspondence: subgroups with more similar attribution patterns tended to exhibit more comparable model performance, whereas subgroups with more divergent feature representations often showed larger gaps in predictive accuracy. For example, in ICU mortality prediction for Hospital 73, the full test set and Subgroup A shared the exact same top five feature representations (Figs. 3a, 3f), with only a modest 5.99% increase in median AUPRC and no statistically significant difference. By contrast, the full test set and Subgroup B for the same hospital displayed notably different attribution patterns (Figs. 3a, 3k), accompanied by a substantial and statistically significant 67.44% increase in median AUPRC. Similar trends were also observed when comparing subgroups with their respective retrained models.

**Fig. 3** SHAP values of the top five most important variables across subgroups, datasets and prediction outcomes. **a–e**: all test group; **f–j**: Subgroup A; **k–o**: Subgroup B; **p–t**: Subgroup A retrained; **u–y**: Subgroup B retrained. These correspond respectively to ICU mortality prediction on hospitals 73, 458, and 420 from the eICU dataset; in-hospital death prediction on hospital 19 from the WiDS dataset; and hidden hypoxemia prediction on the MIMIC IV dataset.

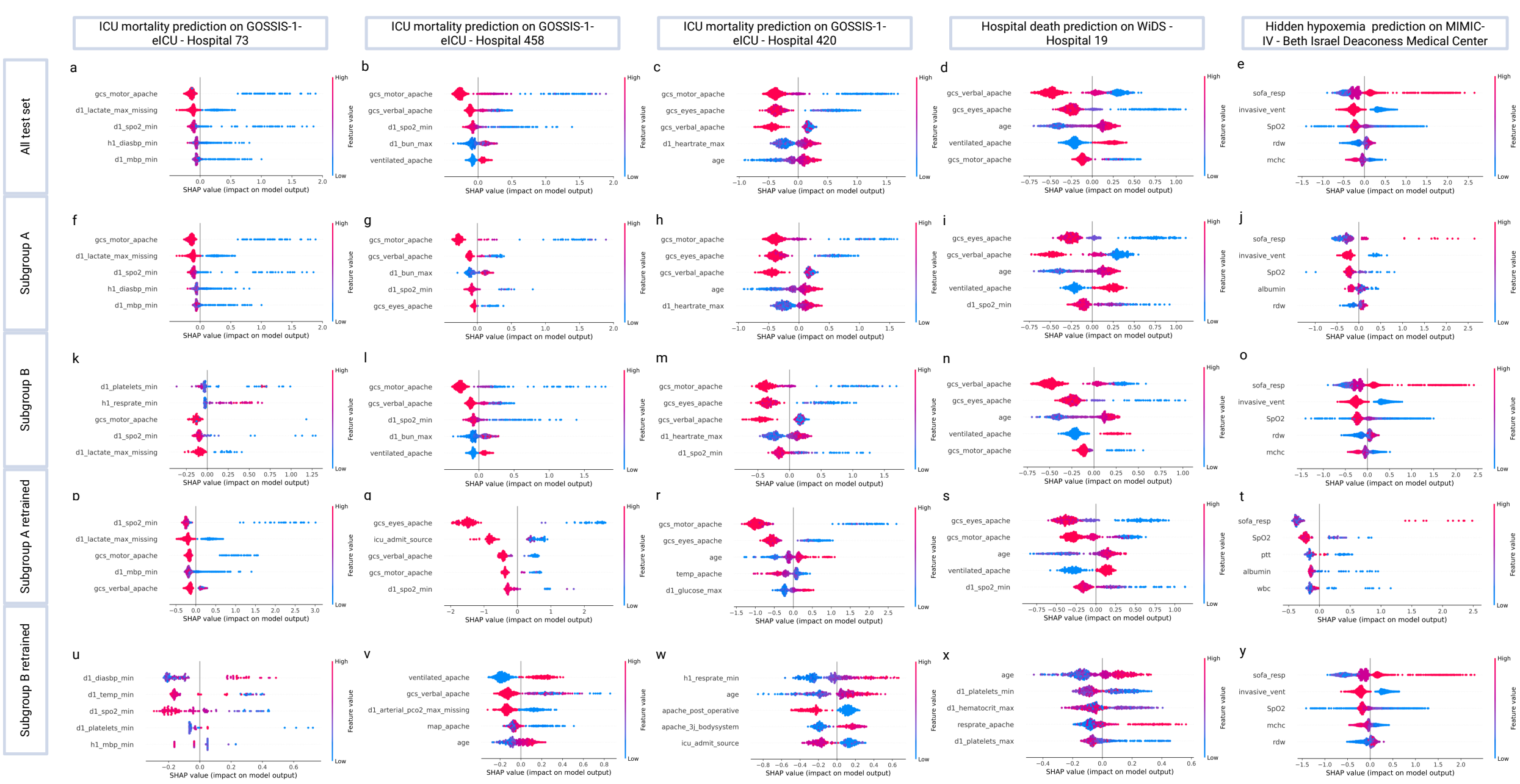

In summary, our findings in this section demonstrate a strong correspondence between SHAP-based feature representations and subgroup-specific model performance across multiple datasets and clinical outcomes. This alignment validates the efficacy of our closed-loop AdaptHetero framework in uncovering clinically meaningful subpopulations and enabling targeted improvements in predictive accuracy. By simultaneously highlighting potential risks, AdaptHetero holds promise for personalized and context-aware deployment of ML models in high-stakes clinical settings.

## Discussion

Heterogeneity of EHR data can introduce cohort-specific biases and noise, potentially degrading model performance and limiting deployment and generalizability.[16,17,26] At the same time, ML models often benefit from training on larger datasets.[27] Ideally, optimal performance is achieved when leveraging a cohort that is both large and relatively homogeneous—balancing representativeness with reduced heterogeneity.[26,27] To this end, we propose AdaptHetero.

AdaptHetero is a novel, closed-loop computational framework designed to identify clinically meaningful, subgroup-specific characteristics. It improves predictive performance across diverse clinical outcome predictions—with gains of up to 174.39%—while also flagging potential risks in others.

To demonstrate its practical utility, consider a clinical deployment scenario where a new ICU patient is admitted. AdaptHetero first calculates the patient's SHAP values to assign them to the most relevant subgroup, thereby identifying potential subgroup-specific risks. It then generates optimized, subgroup-aware risk assessments accompanied by uncertainty estimates and feature representations to support tailored clinical decision-making (see Fig. 4).

Fig. 4 AdaptHetero clinical deployment workflow for subgroup-specific risk prediction and decision support.

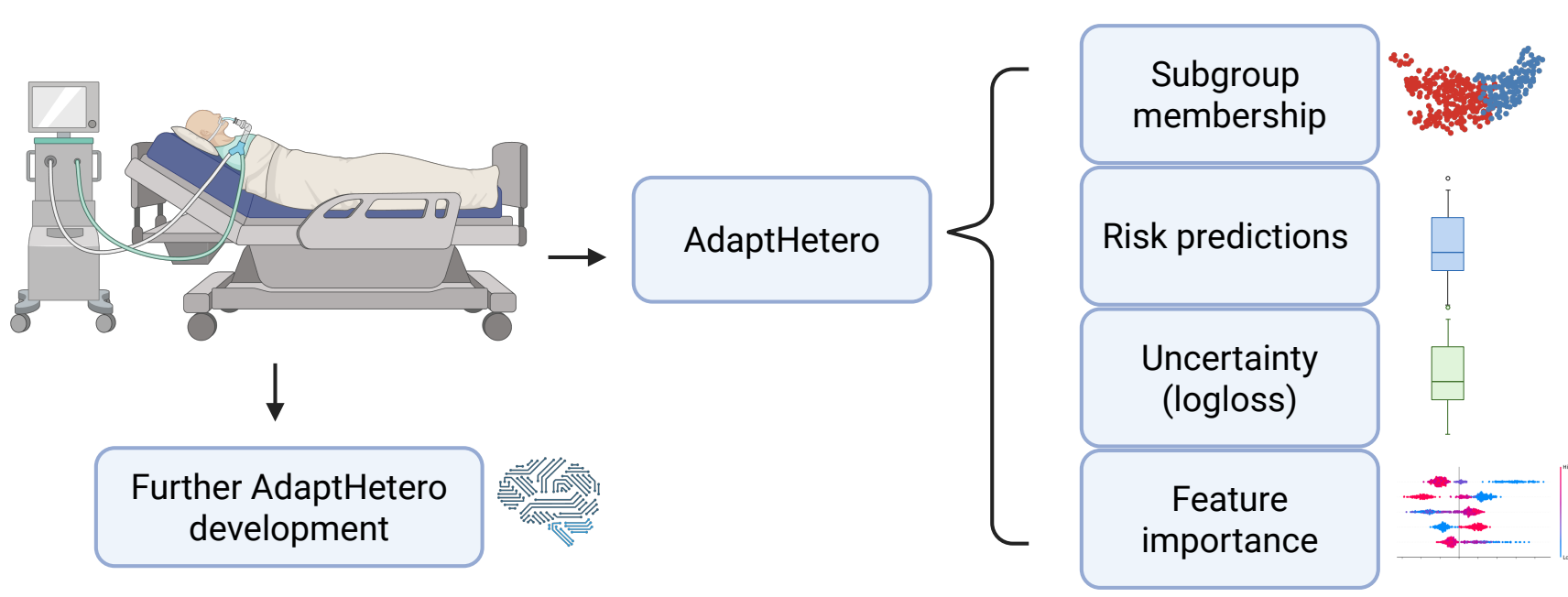

Our results highlight several additional strengths of AdaptHetero: (i) it flexibly accommodates variations in the types and numbers of input variables selected independently by each model development process (see Supplementary Table 2); (ii) it reduces reliance on extensive external validation datasets by operationalizing internal heterogeneity signals, mitigating challenges related to data access and privacy; and (iii) it achieves these benefits using lightweight, interpretable base models that require substantially less computational overhead than deep learning models, facilitating easier deployment and broader applicability in real-world clinical settings.

Previous studies mostly focused on either enhancing overall model robustness through data-centric approaches (e.g., federated learning, data harmonization) or improving generalizability via model-centric strategies such as domain adaptation and representation learning. While these approaches have yielded progress, they often require extensive access to external data sources, incur high computational costs, and struggle to capture the nuanced, latent heterogeneity within and across healthcare institutions.

For instance, federated learning enables collaborative model training without original data sharing, yet it often suffers from covariate shift—such as demographic, clinical, and other heterogeneities—and inconsistent feature semantics across sites.[1] Similarly, deep learning-based domain generalization methods aim to learn transferable representations, but they typically require large-scale, well-labeled datasets and lack theoretically supported interpretability—posing barriers to clinical adoption.[14,26,28]

Moreover, many prior efforts treat heterogeneity as noise to be minimized, rather than a source of potentially meaningful clinical variation. This perspective risks overlooking subgroup-specific signals that could be crucial for personalized care or local decision support.[6,7,29]

Additionally, in EHRs, missing data are often not random artifacts but can reflect underlying clinical behavior or patient condition—for example, the absence of a test order might reflect either lower or higher perceived clinical risk, or institutional resource constraints (Figs. 3a, 3f, 3k, 3p, 3v).

Attempts to stratify patients using clinical rules or phenotype-based clustering have shown promise, but these methods often operate independently of the model learning process and may not reflect the actual feature attribution dynamics that influence predictions.[27,30]

There are several limitations to our study. First, the retrospective nature and reliance on publicly available datasets may introduce biases. Second, although we identified patient subgroups, small sample sizes for some subpopulations limited the ability to train models individually. For such groups with limited data or higher heterogeneity, future work will focus on improving analytical capabilities through additional data collection and integrating complementary approaches such as federated learning.

## Methods

### Datasets

We applied our framework to three datasets,  the eICU-CRD subset of the Global Open Source Severity of Illness Score dataset(GOSSIS-1-eICU), Women in Data Science(WiDS) Datathon 2020: ICU Mortality Prediction, and MIT Critical Datathon 2023: a MIMIC-IV Derived Dataset for Pulse Oximetry Correction Models.  The GOSSIS-1-eICU dataset, comprising 131,051 unique ICU admissions from 204 hospitals during 2014–2015, was curated by excluding readmissions, patients under 16 years of age, and those with missing outcome data or no recorded heart rate. We then selected the top 5 hospitals with most patients: Hospitals 73, 264, 338, 420, and 458. For WiDS Datasets which contains 91,713 labeled encounters spanning more than 130,000 hospital ICU visits from patients, spanning a one-year timeframe, we extracted the top 3 hospitals with most patients: Hospitals 19, 118, and 188. For the MIMIC-IV Derived Dataset for Pulse Oximetry Correction Models that includes de-identified EHR data from 50,920 unique patients at Beth Israel Deaconess Medical Center, in Boston, MA, between 2008 - 2019, we randomly selected one record per patient to avoid data leakage and ensure independence across samples. Detailed patient size and positive rate included for model development are provided in Supplementary Table 1.

**Data preprocessing, processing and imputation**

Data processing was designed to retain critical information for model development while rigorously preventing data leakage between training and test sets. For the GOSSIS-1-eICU data, a total of 20 variables—including patient id, ICU id, hospital bed size, hospital death, APSIII score, and other features deemed irrelevant, duplicated, or at risk of introducing data leakage—were first removed. The variables ultimately included for model development can be found in Supplementary Table 1.

Let the preprocessed dataset be denoted by $\mathcal{D}_{\mathrm{raw}}$, and the cleaned dataset as $\mathcal{D} \subset \mathcal{D}_{\mathrm{raw}}$. The processing steps to generate $\mathcal{D}$ were summarized as follows:

1. If applicable, GCS adjustment:

For the Glasgow Coma Scale features $\{\mathrm{gcs_motor}, \mathrm{gcs_verbal}, \mathrm{gcs_eyes}\}$, if $\mathrm{gcs_unable_apache} = 1$, then we set:

$$\mathrm{gcs_motor} = \mathrm{gcs_verbal} = \mathrm{gcs_eyes} = 0.$$

2. Category completion:

When the features apache_2_bodysystem or apache_3j_bodysystem was missing, they were filled with the value Unavailable.

3. Missingness indicators:

For every feature $x_j \in \mathcal{D}$, $x_j \neq \mathrm{icu_death}$, a binary missingness indicator was created as:

$$m_j^{(i)} = \begin{cases} 1, & \text{if } x_j^{(i)} \text{ is missing} \\ 0, & \text{otherwise.} \end{cases}$$

These indicators were appended to the dataset as new features.

4. Train-test splitting:

The dataset was stratified based on the binary target variable $y = \mathrm{icu_death}$, and partitioned into training and testing sets:

$$\mathcal{D}_{\mathrm{train}}, \mathcal{D}_{\mathrm{test}} = \mathrm{StratifiedSplit}(\mathcal{D}, \mathrm{ratio} = 0.7{:}\,0.3)$$

5. Feature exclusion by missingness:

Define the missing rate of feature $x_j$ as:

$$r_j = \frac{1}{n_{\mathrm{train}}} \sum_{i=1}^{n_{\mathrm{train}}} \delta(x_j^{(i)} \text{ is missing})$$

Features with $r_j > 0.10$ were excluded from both training and test sets.

6. Feature selection by statistical association:

Let $x_j$ be a candidate predictor in training set. If $x_j$ is categorical, its association with y was evaluated by the Chi-squared test:

$$p_j = \mathcal{X}^2(x_j, y), \qquad \text{retain if } p_j < 0.05$$

If $x_j$ is continuous, the Mann-Whitney U test was used:

$$p_j = \mathrm{MWU}(x_j \mid y = 0, x_j \mid y = 1), \qquad \text{retain if } p_j < 0.05$$

Only variables that passed this statistical association test in the training set were retained; all other variables were removed from both the training and test sets to ensure consistency.

7. Imputation:
Remaining missing values were imputed as:
1) for continuous features:

$$x_j^{(i)} = \mathrm{median}(x_j), \text{if } x_j^{(i)} \text{ is missing}$$

2) for categorical features:

$$x_j^{(i)} = \mathrm{mode}(x_j), \text{if } x_j^{(i)} \text{ is missing}$$

Median and mode values were computed separately within the training and test sets.
8. Categorical encoding:
Let Cat $\subset \mathcal{D}_{\mathrm{train}}$ be the set of categorical variables. An ordinal encoder $\mathcal{O}(\cdot)$was trained on $\mathrm{Cat}_{\mathrm{train}}$ and applied to both training and testing set:

$$\mathrm{Cat}_{\mathrm{train}}^{\mathrm{enc}} = \mathcal{O}(\mathrm{Cat}_{\mathrm{train}}), \qquad \mathrm{Cat}_{\mathrm{test}}^{\mathrm{enc}} = \mathcal{O}(\mathrm{Cat}_{\mathrm{test}})$$

9. Reindexing:
Final datasets were reindexed so that all samples have unique, consecutive integer indices:

$$\mathrm{index}_{x_i} = i, i = 0, 1, \dots, n-1$$

They were prepared for accurate subgroup mapping and downstream analyses.
For the WiDS dataset, we removed patient ID, ICU and hospital discharge location, encounter ID, APSIII score, and 13 additional features deemed irrelevant, redundant, or posing a risk of data leakage. The prediction target for this dataset was hospital_death. Subsequent processing steps followed the same procedures as applied to the GOSSIS-1-eICU dataset.
For the MIMIC-IV Derived Dataset for Pulse Oximetry Correction Models, instances where the time offset of peripheral oxygen saturation measurements (delta_SpO2) exceeded 60 minutes were excluded. Additionally, SaO2, gender, all delta-related variables, and 9 other features identified as irrelevant, redundant, or potential sources of data leakage were removed. The prediction target for this dataset was hidden_hypoxemia. All remaining processing steps were consistent with those used for the GOSSIS-1-eICU dataset.

**AdaptHetero architecture**
To identify clinically meaningful patient subgroups with model-derived interpretation, AdaptHetero framework implemented a hybrid supervised-unsupervised learning paradigm combining model training, SHAP representation, dimensionality reduction, clustering, and label transforming.
For processed hospital data, a predictive model, $f_{\mathrm{model}}$ was constructed using XGBoost. The training dataset $\mathcal{D}_{\mathrm{train}}$ was split into two partitions: 80% was used to train an initial model, and the remaining was reserved for hyperparameter tuning. Once the optimal parameters were identified, the final model $f_{\mathrm{model}}$ was retrained on the full training dataset $\mathcal{D}_{\mathrm{train}}$.
Given the trained model $f_{\mathrm{model}}$, SHAP values were computed for both the training and testing datasets as $\Phi_{\mathrm{train}} = \Phi(\mathcal{D}_{\mathrm{train}})$ and $\Phi_{\mathrm{test}} = \Phi(\mathcal{D}_{\mathrm{test}})$, where $\Phi(\cdot)$ denotes the SHapley Additive exPlanations function. SHAP values were then standardized using a normalization function $S(\cdot)$, yielding $\widetilde{\Phi}_{\text{train}}$ and $\widetilde{\Phi}_{\text{test}}$. To visualize and cluster samples in a lower-dimensional space, UMAP was applied as a projection $U(\cdot): \mathbb{R}^d \rightarrow \mathbb{R}^2$, resulting in embeddings:

$$\mathcal{E}_{\text{train}} = U(\widetilde{\Phi}_{\text{train}}),\ \mathcal{E}_{\text{test}} = U(\widetilde{\Phi}_{\text{test}})$$

Clustering was performed on the training embeddings using HDBSCAN, defined as $C_{\text{train}} = C(\mathcal{E}_{\text{train}})$. These cluster were then extended to the test set, KNN algorithm was utilized to assign each test sample to the most common label among its nearest neighbors in $\mathcal{E}_{\text{train}}$, resulting in cluster labels:

$$C_{\text{test}} = K(\mathcal{E}_{\text{test}} \,|\mathcal{E}_{\text{train}}, C_{\text{train}}).$$

With the obtained cluster labels, the trained model $f_{\text{model}}$ was evaluated on all possible subgroup combinations derived from $C_{\text{train}}$ and $C_{\text{test}}$ . Let M denote the number of unique clusters identified in the training set $\mathcal{D}_{\text{train}}$, i.e., $K = |\{c \in C_{\text{train}}\}|$. The total number of non-empty cluster combinations, potential subgroups $\mathcal{S}$, is then given by:

$$N_{\text{subgroups}} = 2^M - 1$$

For each subgroup $\mathcal{S} \subseteq \mathcal{D}_{\text{train}} \cup \mathcal{D}_{\text{test}}$, SHAP values were recalculated to interpret local feature contributions. Based on the cluster wise sample size, model predictive metrics and interpretability of SHAP patterns, either zero or two subgroups $\mathcal{S}_{\text{sel}}$ were selected for further remodeling. Specifically, Hospitals 264, 338, 118, and 188 were excluded from further remodeling (e.g., Hospital 338 was excluded because it had only three clusters, two of which contained a total of 107 samples in the training set). A new model $f_{\text{sel}}$ for every subgroup was then trained and tuned using the same procedure as described above, but with restricted to the training data within $\mathcal{S}_{\text{sel}}$, and subsequently evaluated on the corresponding test subset $\mathcal{S}_{\text{sel}}^{\text{test}}$.

**Data availability:** The data utilized in this study is available at https://physionet.org/. Access to these resource are restricted; proper registration, required training, and signed data usage agreements may be required.

**Code availability**: The code developed in this study is available at https://github.com/lingliao/Interpretable_AL

**Acknowledgements** L.L. would like to acknowledge the consistent support from McDonnell International Scholars Academy at Washington University in Saint Louis.

**Author contributions** L.L. and E.A. conceptualized and designed the study. L.L. was responsible for the acquisition, analysis, and interpretation of the data. L.L. and E.A. drafted the manuscript. L.L. and E.A. critically reviewed the manuscript for important intellectual content. L.L. performed the statistical analysis. L.L. and E.A. supervised the study. All authors reviewed the manuscript.

**Competing interests** L.L. is the founder of Biomedical Deep Learning LLC. E.A. has no conflict of interest to disclosure.

**References**

[1] Zhu, H., Bai, J., Li, N. et al. FedWeight: mitigating covariate shift of federated learning on electronic health records data through patients re-weighting. npj Digit. Med. 8, 286 (2025). https://doi.org/10.1038/s41746-025-01661-8

[2] Liu, X., Delahanty, R. J., Engelberg, R. A., Reineck, L. A., Ratcliffe, S. J., Asch, D. A., ... & Halpern, S. D. (2023). Illness severity assessment of older adults in critical illness using machine learning (ELDER-ICU): An international multicentre study with subgroup bias evaluation. The Lancet Digital Health, 5(10), e657–e667. https://doi.org/10.1016/S2589-7500(23)00140-4

[3] Murdoch, W. J., Singh, C., Kumbier, K., Abbasi-Asl, R., & Yu, B. (2019). Definitions, methods, and applications in interpretable machine learning. Proceedings of the National Academy of Sciences of the United States of America, 116(44), 22071–22080. https://doi.org/10.1073/pnas.1900654116

[4] Lundberg, S. M., & Lee, S.-I. (2017). A unified approach to interpreting model predictions. Advances in Neural Information Processing Systems, 30. https://proceedings.neurips.cc/paper_files/paper/2017/file/8a20a8621978632d76c43dfd28b67767-Paper.pdf

[5] Liang K. Y. (1987). Extended Mantel-Haenszel estimating procedure for multivariate logistic regression models. Biometrics, 43(2), 289–299.

[6] Wu, HD.I. (2004). Effect of Ignoring Heterogeneity in Hazards Regression. In: Balakrishnan, N., Nikulin, M.S., Mesbah, M., Limnios, N. (eds) Parametric and Semiparametric Models with Applications to Reliability, Survival Analysis, and Quality of Life. Statistics for Industry and Technology. Birkhäuser, Boston, MA. https://doi.org/10.1007/978-0-8176-8206-4_16

[7] Tong, J., Duan, R., Li, R., Scheuemie, M. J., Moore, J. H., & Chen, Y. (2020). Robust-ODAL: Learning from heterogeneous health systems without sharing patient-level data. Pacific Symposium on Biocomputing. Pacific Symposium on Biocomputing, 25, 695–706.

[8] McMahan, H. B., Moore, E., Ramage, D., Hampson, S. & Aguera y Arcas, B. Communication-efficient learning of deep networks from decentralized data. Proceedings of the 20th International Conference on Artificial Intelligence and Statistics (AISTATS) (2017).

[9] Shimodaira, H. Improving predictive inference under covariate shift by weighting the log-likelihood function. J. Stat. Plan. Inference 90, 227–244 (2000).

[10] Valan, B., Prakash, A., Ratliff, W. et al. Evaluating sepsis watch generalizability through multisite external validation of a sepsis machine learning model. npj Digit. Med. 8, 350 (2025). https://doi.org/10.1038/s41746-025-01664-5

[11] Chalkidis G, McPherson JP, Beck A, et al. External Validation of a Machine Learning Model to Predict 6-Month Mortality for Patients With Advanced Solid Tumors. JAMA Netw Open. 2023;6(8):e2327193. doi:10.1001/jamanetworkopen.2023.27193

[12] Lou SS, Kumar S, Goss CW, et al. Multicenter Validation of a Machine Learning Model for Surgical Transfusion Risk at 45 US Hospitals. JAMA Netw Open. 2025;8(6):e2517760. doi:10.1001/jamanetworkopen.2025.17760

[13] Raffa, J. D., Johnson, A. E. W., O'Brien, Z., Pollard, T. J., Mark, R. G., Celi, L. A., Pilcher, D., & Badawi, O. (2022). The Global Open Source Severity of Illness Score (GOSSIS). Critical care medicine, 50(7), 1040–1050. https://doi.org/10.1097/CCM.0000000000005518

[14] Amirahmadi, A., Ohlsson, M., & Etminani, K. (2023). Deep learning prediction models based on EHR trajectories: A systematic review. Journal of biomedical informatics, 144, 104430. https://doi.org/10.1016/j.jbi.2023.104430

[15] Saeed, N., Ashour, M. & Mashaly, M. Comprehensive review of federated learning challenges: a data preparation viewpoint. J Big Data 12, 153 (2025). https://doi.org/10.1186/s40537-025-01195-6

[16] Miotto, R., Wang, F., Wang, S., Jiang, X. & Dudley, J. T. Deep learning for healthcare: Review, opportunities and challenges. Briefings Bioinf. 19, 6 (2018).

[17] Ghassemi, M., Naumann, T., Schulam, P., Beam, A. L., Chen, I. Y., & Ranganath, R. (2020). A Review of Challenges and Opportunities in Machine Learning for Health. AMIA Joint Summits on Translational Science proceedings. AMIA Joint Summits on Translational Science, 2020, 191–200.

[18] Scott M. Lundberg and Su-In Lee. 2017. A unified approach to interpreting model predictions. In Advances in Neural Information Processing Systems. 4765–4774.

[19] Lee, M., Raffa, J., Ghassemi, M., Pollard, T., Kalanidhi, S., Badawi, O., Matthys, K., & Celi, L. A. (2020). WiDS (Women in Data Science) Datathon 2020: ICU Mortality Prediction (version 1.0.0). PhysioNet. RRID:SCR_007345. https://doi.org/10.13026/vc0e-th79

[20] Matos, J., Struja, T., Restrepo, D. S., Nakayama, L. F., Gallifant, J., Weishaupt, L., Mullangi, N., Loureiro, M., Shapiro, S., Carrel, A., & Celi, L. A. (2023). MIT Critical Datathon 2023: a MIMIC-IV Derived Dataset for Pulse Oximetry Correction Models (version 1.0.0). PhysioNet. RRID:SCR_007345. https://doi.org/10.13026/jfpc-pz79

[21] McInnes et al., (2017). hdbscan: Hierarchical density based clustering. Journal of Open Source Software, 2(11), 205, doi:10.21105/joss.00205

[22] Bampa, M., Miliou, I., Jovanovic, B., & Papapetrou, P. (2024). M-ClustEHR: A multimodal clustering approach for electronic health records. Artificial intelligence in medicine, 154, 102905. https://doi.org/10.1016/j.artmed.2024.102905

[23] Guo, G., Wang, H., Bell, D., Bi, Y., Greer, K. (2003). KNN Model-Based Approach in Classification. In: Meersman, R., Tari, Z., Schmidt, D.C. (eds) On The Move to Meaningful Internet Systems 2003: CoopIS, DOA, and ODBASE. OTM 2003. Lecture Notes in Computer Science, vol 2888. Springer, Berlin, Heidelberg. https://doi.org/10.1007/978-3-540-39964-3_62

[24] Aljohani, A. Optimizing Patient Stratification in Healthcare: A Comparative Analysis of Clustering Algorithms for EHR Data. Int J Comput Intell Syst 17, 173 (2024). https://doi.org/10.1007/s44196-024-00568-8

[25] Nachar, N. 2008. “The Mann-Whitney U: A Test for Assessing whether Two Independent Samples Come from the Same Distribution.” Tutorials in Quantitative Methods for Psychology 4: 13–20.

[26] Burns, M. L., & Kheterpal, S. (2020). Machine Learning Comes of Age: Local Impact versus National Generalizability. Anesthesiology, 132(5), 939–941. https://doi.org/10.1097/ALN.0000000000003223

[27] Rajkomar, A., Dean, J., & Kohane, I. (2019). Machine learning in medicine. New England Journal of Medicine, 380(14), 1347–1358. https://doi.org/10.1056/NEJMra1814259

[28] Landi, I., Glicksberg, B.S., Lee, HC. et al. Deep representation learning of electronic health records to unlock patient stratification at scale. npj Digit. Med. 3, 96 (2020). https://doi.org/10.1038/s41746-020-0301-z

[29] Collins, F. S., & Varmus, H. (2015). A new initiative on precision medicine. The New England journal of medicine, 372(9), 793–795. https://doi.org/10.1056/NEJMp1500523

[30] Fan, Z., et al. (2024). Identification of heart failure subtypes using transformer-based deep learning modelling: A population-based study of 379,108 individuals. eBioMedicine, 114, 105657. https://doi.org/10.1016/j.ebiom.2025.105657